\documentclass[letterpaper, 10 pt, conference]{IEEEtran}
\IEEEoverridecommandlockouts

\usepackage{times}
\usepackage{makecell}

\usepackage{etoolbox}
\makeatletter
\patchcmd{\@makecaption}
  {\scshape}
  {}
  {}
  {}
\makeatother

\usepackage{tikz}

\usepackage[numbers]{natbib}
\usepackage{multicol}
\usepackage{multirow}
\usepackage{graphicx}
\usepackage[bookmarks=true]{hyperref}
\usepackage{amsmath}
\usepackage{hyphenat}
\usepackage{amssymb}
\usepackage{float}
\usepackage{pifont}
\usepackage{placeins}
\usepackage{mathdots}
\usepackage[caption=false,font=normalsize,labelfont=sf,textfont=sf]{subfig}
\usepackage{xspace}
\usepackage{svg}
\usepackage{rotating}
\usepackage{fontawesome}
\usepackage{pifont}

\begin{document}

\title{ \LARGE \bf
RadarSplat-RIO: Indoor Radar-Inertial Odometry with Gaussian Splatting-Based Radar Bundle Adjustment}

\author{Pou-Chun Kung$^{1,2,*}$, Yuan Tian$^{1}$, Zhengqin Li$^{1}$, Yue Liu$^{1}$, Eric Whitmire$^{1}$, Wolf Kienzle$^{1}$, and  Hrvoje Benko$^{1}$
\thanks{*Work done during his internship at Meta.}
\thanks{$^{1}$P.C. Kung, Y. Tian, Z. Li, Y. Liu, E Whitmire, W. Kienzle, H. Benko were with Meta Reality Labs, Redmond, WA 98052 at the time this work was completed
        }
\thanks{$^{2}$P.C. Kung is with the Department of Robotics, University of Michigan, Ann Arbor, MI 48109 {\tt\small pckung@umich.edu}
}
}


\maketitle
\begin{abstract}
Radar is more resilient to adverse weather and lighting conditions than visual and Lidar simultaneous localization and mapping (SLAM). However, most radar SLAM pipelines still rely heavily on frame-to-frame odometry, which leads to substantial drift.
While loop closure can correct long-term errors, it requires revisiting places and relies on robust place recognition. In contrast, visual odometry methods typically leverage bundle adjustment (BA) to jointly optimize poses and map within a local window. However, an equivalent BA formulation for radar has remained largely unexplored.
We present the first radar BA framework enabled by Gaussian Splatting (GS), a dense and differentiable scene representation. Our method jointly optimizes radar sensor poses and scene geometry using full range-azimuth-Doppler data, bringing the benefits of multi-frame BA to radar for the first time. When integrated with an existing radar-inertial odometry frontend, our approach significantly reduces pose drift and improves robustness.
Across multiple indoor scenes, our radar BA achieves substantial gains over the prior radar-inertial odometry, reducing average absolute translational and rotational errors by $\sim$90\% and $\sim$80\%, respectively.

\end{abstract}

\begin{figure}[t!]
    \centering
    \includegraphics[width=0.99\linewidth]{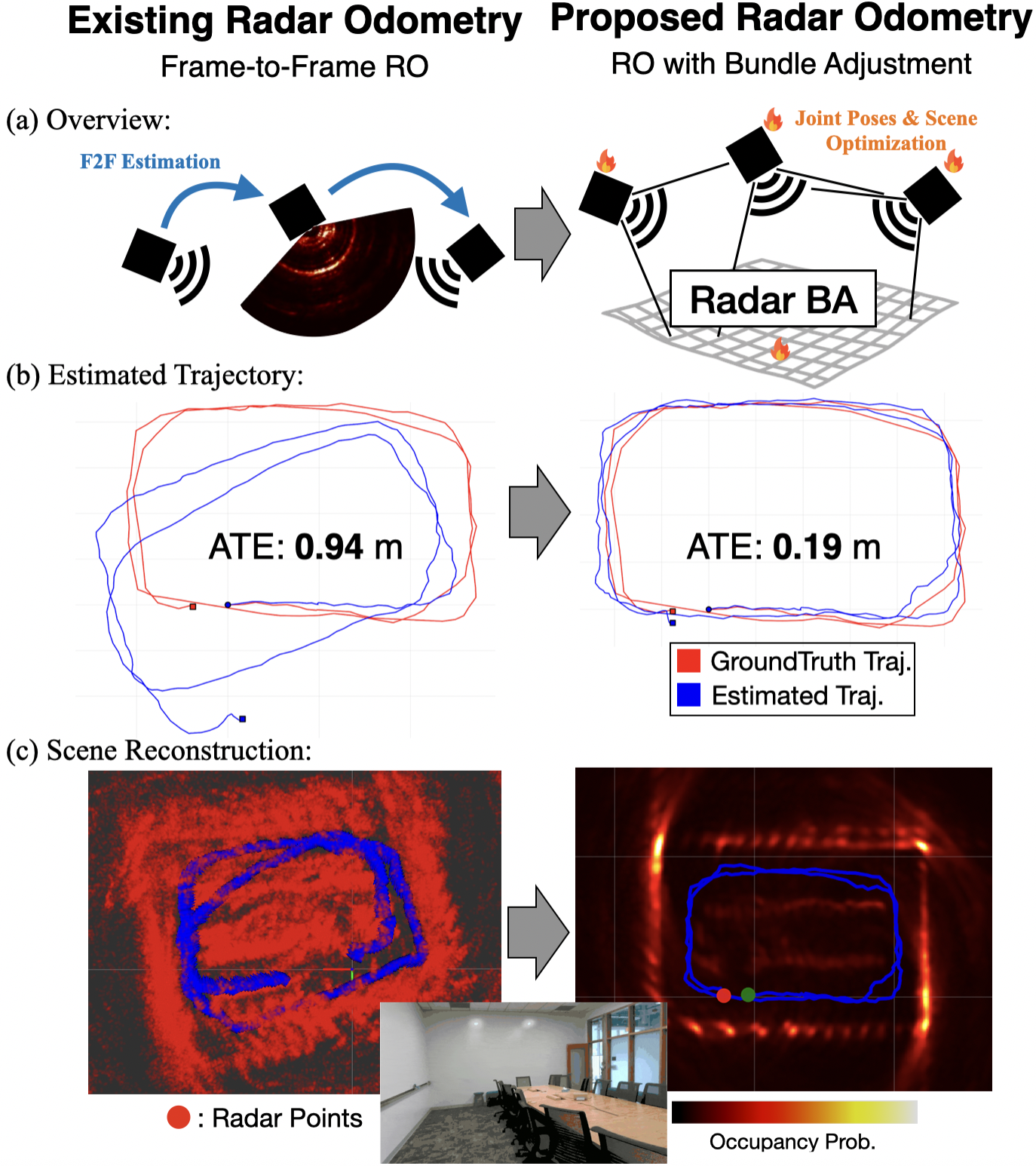}
        \vspace{-0.2in}
        \caption{Existing radar SLAM methods largely depend on frame-to-frame estimation, which inevitably accumulates drift over time. To address this limitation, we introduce a Gaussian Splatting–based radar bundle adjustment framework that jointly optimizes multi-frame poses and a dense scene representation. This joint optimization substantially reduces drift and improves overall trajectory consistency. 
        }
        \vspace{-0.15in}
        \label{fig:teaser_fig}
\end{figure}

\section{Introduction}
Simultaneous Localization and Mapping (SLAM) is a fundamental capability for autonomous systems. While most SLAM systems rely on optical sensors such as cameras and LiDAR, these modalities suffer from performance degradation under adverse weather conditions (e.g., rain, fog, snow) or variable lighting environments (e.g., low light or overexposure).
Millimeter-wave radar offers a compelling alternative due to its robustness to illumination changes~\cite{radarlowlight}, harsh weather~\cite{radiate, radarslam}, and occlusions by non-metallic materials (e.g., plastic, fabric)~\cite{radarpenetration}. 
Moreover, the radar is low-cost and has low power consumption.
These properties make radar particularly suitable for challenging SLAM applications in autonomous robotics.

A wide range of radar-based odometry methods have been proposed, including velocity-based~\cite{radar_ego_vel}, geometry-based~\cite{ndt_ro, cen_ro, pharao}, and learning-based approaches~\cite{masking_by_moving, radarize}. However, most radar and radar-inertial odometry (RO/RIO) methods estimate poses by integrating estimated per-frame velocities or matching radar points or images between consecutive frames, effectively following a dead-reckoning paradigm~\cite{mobilerobot}. Consequently, drift accumulates over time. Although some methods incorporate loop closure and place recognition~\cite{spr, raislam, scancontext, rivslam} to counteract this drift, such corrections require revisiting previously observed locations, limiting their applicability in online or exploratory applications.

\begin{figure}[t!]
    \centering
    \includegraphics[width=0.8\linewidth]{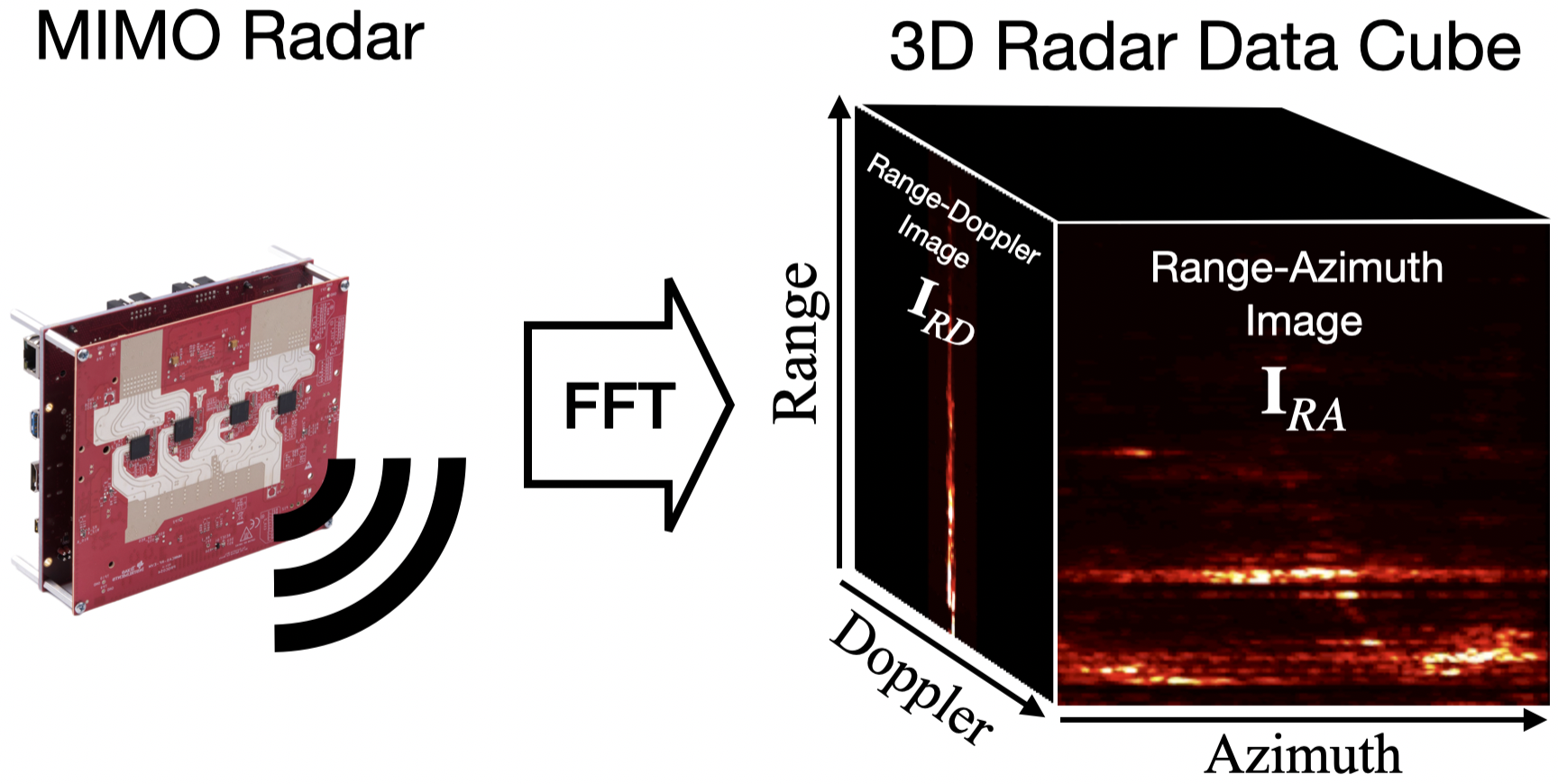}
        \vspace{-0.05in}
        \caption{Illustration of 3D radar data from a Multiple-Input and Multiple-Output (MIMO) radar after FFT-based signal processing. The resulting data cube spans three dimensions corresponding to range, azimuth, and Doppler velocity measurements.}
        \vspace{-0.15in}
        \label{fig:radar_data}
\end{figure}

Multi-frame bundle adjustment (BA) is a standard component of visual odometry (VO) systems for reducing drift. It jointly optimizes camera poses and scene structure within a sliding window~\cite{orbslam2, vinsmono}. This optimization significantly improves VO accuracy without relying on loop closures or revisiting previously seen locations. 
Recent work has shown that LiDAR can also benefit from BA-style optimization~\cite{balm, loner, pinslam, splatloam}. 
However, adapting BA to radar remains largely unexplored and challenging due to the absence of dense scene representation and the complexity of radar’s sensing modality, which involves range, azimuth, and Doppler dimensions, as shown in Figure~\ref{fig:radar_data}.

To address this gap, we introduce \textbf{RadarSplat-RIO}, the first radar BA framework inspired by RadarSplat~\cite{radarsplat}, a dense and differentiable Gaussian Splatting (GS)~\cite{3dgs} representation for radar. 
Prior NeRF~\cite{nerf}- and GS-based visual and LiDAR SLAMs~\cite{gsslam, niceslam, loner} have shown that BA with dense representation improves pose estimation. We aim to extend this paradigm to radar SLAM.
However, while RadarSplat supports radar rendering, it does not model Doppler velocity measurements.
To fully exploit radar sensing capability, we propose \textbf{RadarSplat++}, which augments RadarSplat with Doppler-aware rendering inspired by DART~\cite{dart}, a radar NeRF method that supports range-Doppler (RD) rendering. By leveraging GS-based BA and full radar range-azimuth-Doppler (RAD) measurement, RadarSplat-RIO achieves accurate and drift-resilient radar odometry without loop closures.
Our key contributions are:

\begin{itemize}
\item We present the first \textbf{radar bundle adjustment} framework that jointly optimizes radar sensor poses and scene geometry within a Gaussian Splatting representation.
\item We propose \textbf{RadarSplat++}, the first radar rendering pipeline that supports full radar measurements, including range, azimuth, and Doppler dimensions.
\item We demonstrate \textbf{significant improvements in RIO} by integrating our BA backend with existing RIO frontends, achieving reduced drift and enhanced robustness over long-term trajectories.
\end{itemize}

\section{Related Work}

\subsection{Radar-based SLAM/Odometry}
Radar odometry and SLAM can be approached in various ways depending on the radar sensor type and the provided data format. In general, existing radar SLAM methods can be categorized into velocity-based, geometry-based, learning-based, or their combinations.

Velocity-based methods are the most classical and widely adopted radar odometry~\cite{radar_ego_vel, ekf_rio, xrio, mrio}. These approaches estimate the ego-velocity directly from Doppler measurements by assuming that most observed targets are static. However, relying solely on velocity integration inevitably leads to drift when the velocity estimates contain even small errors.
Geometry-based methods have leveraged geometric information in radar images or points for pose estimation. These approaches estimate poses by associating radar points across frames~\cite{ndt_ro, cen_ro, orora, perio,rai_slam} or matching consecutive radar images~\cite{pharao, wicp_ro, randt_slam}. 
Although geometric constraints improve stability, frame-to-frame estimation still accumulates error over long trajectories.
Finally, learning-based approaches have been proposed for feedforward pose or velocity prediction from radar images~\cite{masking_by_moving, radarize} or points~\cite{equi_ro}. These methods remain per-frame estimation and thus still suffer from error accumulation.

Radar–inertial SLAM systems typically combine radar-estimated translational velocity with IMU angular velocity for accurate pose estimation. Early methods~\cite{ekf_rio, xrio} employ filtering-based fusion, whereas optimization-based formulations are adopted by~\cite{mrio, perio, rai_slam}. Nevertheless, IMU fusion primarily imposes local angular constraints and does not guarantee global trajectory consistency.


\begin{figure}[t!]
    \centering
    \includegraphics[width=0.99\linewidth]{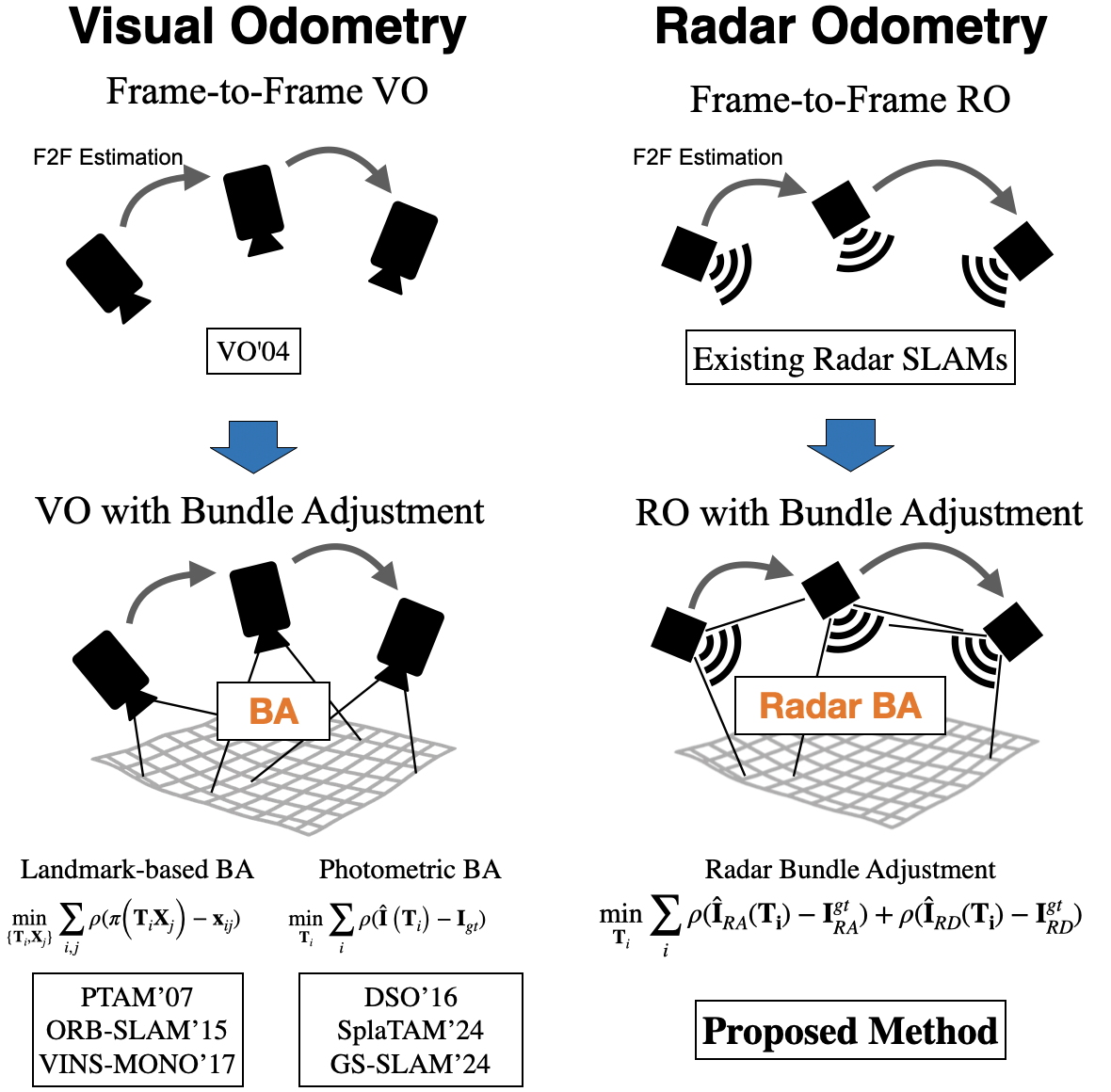}
        \vspace{-0.05in}
        \caption{
        Comparison between visual and radar SLAM. Visual SLAM has progressed from frame-to-frame VO to BA-enabled joint pose–scene optimization. In contrast, existing radar SLAM remains predominantly frame-to-frame. Our method introduces the first radar BA for drift-resilient odometry. $\rho(\cdot)$ can be any loss function.}
        \vspace{-0.1in}
        \label{fig:related_work}
\end{figure}

\subsection{Visual and LiDAR SLAM with Bundle Adjustment}

\subsubsection{Visual Odometry}
Early VO systems relied on frame-to-frame feature matching and pose aggregation~\cite{vo}. PTAM~\cite{ptam} first show that jointly optimizing multi-frame poses and scene structure via BA is essential for preventing drift. Landmark-based BA subsequently became a standard component of VO and visual SLAM~\cite{orbslam2, vinsmono}. More recently, advances in dense scene representations such as NeRF~\cite{nerf} and GS~\cite{3dgs} have enabled photometric BA formulations for NeRF/GS-based SLAM systems~\cite{nerfslam, splatam, gaussianslam, gsslam} with dense scene reconstruction and more accurate pose estimation. The comparison between visual and radar SLAM progress is illustrated in Figure~\ref{fig:related_work}

\subsubsection{LiDAR Odometry}
Similarly, while frame-to-frame or frame-to-map approaches are used in early LiDAR odometry and mapping (LOAM)~\cite{loam}, the BA concept is also adapted to LiDAR in BALM~\cite{balm} with multi-frame poses and map optimization. The idea is also extended to NeRF/GS-based LiDAR SLAM, like~\cite{loner, pinslam, splatloam}.

\subsubsection{Summary}
Despite the success of BA in visual and LiDAR SLAM, applying BA to radar data remains challenging due to the inherent complexity of the radar sensor, unstable feature points, and the lack of a multi-view reconstruction framework, such as NeRF or GS for radar. Inspired by recent advances in RadarSplat~\cite{radarsplat}, we propose a radar bundle adjustment framework built on the same GS–based scene representation. By leveraging the differentiable and continuous nature of GS, our method jointly optimizes radar sensor poses and the underlying scene geometry. This formulation narrows the gap between radar-based SLAM and visual or LiDAR SLAM, leading to improved trajectory accuracy over existing radar odometry methods.








\begin{figure*}[t!]
    \centering
    \includegraphics[width=0.99\linewidth]{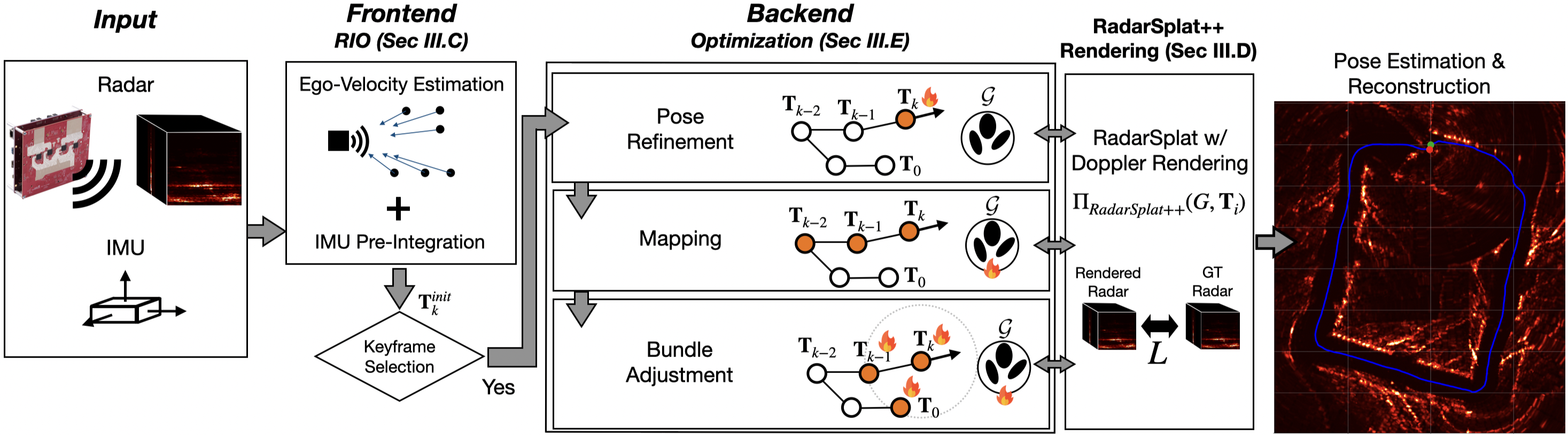}
        \vspace{-0.05in}
        \caption{Pipeline of RadarSplat-RIO. The frontend performs radar–inertial odometry to provide initial pose estimates. The backend refines these poses within a Gaussian Splat representation and maintains a local map using a sliding-window strategy. Finally, the proposed bundle adjustment jointly optimizes radar poses and the scene representation. Orange nodes indicate selected keyframes at different stages. More details is discussed in Sec. ~\ref{sec:backend}.}
        \vspace{-0.1in}
        \label{fig:pipeline}
\end{figure*}

\section{Method}

\subsection{Radar Preliminary}
\subsubsection{Radar Sensor Data}
FMCW MIMO radars generate a raw complex-valued 3D radar data cube. After applying Fast Fourier Transform (FFT) over fast-time samples, antenna channels, and slow-time chirps, the radar produces a cube $\mathbf{I}_{RAD} \in \mathbb{C}^{N_r \times N_a \times N_d}$ spanning range, azimuth, and Doppler (Figure~\ref{fig:radar_data}). To improve computational efficiency, the 3D radar data cube is typically reduced to 2D representations along the range–azimuth or range–Doppler dimensions, forming the RA image $\mathbf{I}{RA} \in \mathbb{R}^{N_r \times N_a}$ and the RD image $\mathbf{I}{RD} \in \mathbb{R}^{N_r \times N_d}$. 
Conventional methods, such as CFAR~\cite{cfar}, further extract feature points in RA image to obtain radar point clouds with position and velocity information for downstream applications. 

\subsubsection{Radar Equation}
A radar-azimuth image $\mathbf{I}_{RA}$ is comprised of multiple range-power signals captured from different azimuth angles.
The received power $P_r(n)$ at bin $n$ with range $R_n$ is determined by the radar equation:
\begin{equation}
    P_r(n) = \frac{P_t G^2  \lambda^2 \sigma }{(4\pi)^3 R_n^4 L},
\label{eq:radar_equation}
\end{equation}
where $P_t$ is the radar peak transmit power, $G$ is the antenna gain depending on elevation angle, $\lambda$ is the wavelength, $L$ represents system and propagation loss, and $\sigma$ is the total radar cross-section (RCS) of objects at range $R_n$.

\subsection{RadarSplat-RIO Pipeline}
Figure~\ref{fig:pipeline} illustrates an overview of the proposed RadarSplat-RIO. An existing RIO~\cite{mrio} is used as the frontend to obtain the initial pose at a new timestamp. In the proposed backend optimization, we follow classical NeRF/GS SLAM~\cite{nerfslam, loner, gaussianslam} to perform pose refinement, mapping, and bundle adjustment processes as the backend. The loss is computed as the difference between the RadarSplat++ rendering and the radar image.

\subsection{Frontend: Radar-Inertial Odometry}
\label{sec:rio}
We follow MRIO~\cite{mrio} to deploy a graph-based RIO as our frontend. The radar ego-velocity is estimated from a single radar image following the formulation of \cite{radar_ego_vel}. 
Each radar image produces a set of feature points, extracted using CFAR~\cite{cfar}, denoted as $\mathbf{p}^{i}$, along with their corresponding Doppler velocities $v_{d}^{i}$, where $i$ indexes the feature points.
The unit direction vector describing line-of-sight from the radar to the target for each point is $\mathbf{r}^i = \frac{\mathbf{p}^i}{\lVert \mathbf{p}^i \rVert}$.
Assuming a static environment, the measured Doppler velocity $v_{d}^i$ equals the projection of the radar ego-velocity 
$\mathbf{v}$ onto $\mathbf{r}_r^i$. Therefore 
$v_{d}^i = (\mathbf{r}^i)^\top \mathbf{v}.$

\begin{equation}
\underbrace{
\begin{bmatrix}
v_{d}^1 \\[2pt]
v_{d}^2 \\[2pt]
\vdots \\[2pt]
v_{d}^N
\end{bmatrix}}_{\mathbf{y}_r}
=
\underbrace{
\begin{bmatrix}
r_{x}^1 & r_{y}^1 & r_{z}^1 \\[2pt]
r_{x}^2 & r_{y}^2 & r_{z}^2 \\[2pt]
\vdots & \vdots & \vdots \\[2pt]
r_{x}^N & r_{y}^N & r_{z}^N
\end{bmatrix}}_{\mathbf{H}}
\underbrace{
\begin{bmatrix}
v_{x} \\[2pt]
v_{y} \\[2pt]
v_{z}
\end{bmatrix}}_{\mathbf{v}},
\label{eq:radar_lsq}
\end{equation}
The radar ego-velocity is estimated using a least-squares solution:
\begin{equation}
    \hat{\mathbf{v}}_r = (\mathbf{H}^\top \mathbf{H})^{-1} \mathbf{H}^\top \mathbf{y}_r.
\end{equation}

The IMU pre-integration factor is defined as \cite{mrio} to fuse radar translation velocity and IMU rotation velocity using the GTSAM library~\cite{gtsam}. 

\subsection{RadarSplat++: RadarSplat with Proposed Range-Doppler Rendering}
\label{sec:radarsplat++}

\subsubsection{RadarSplat Recap}
RadarSplat~\cite{radarsplat} represents the scene as a set of 3D Gaussians following~\cite{3dgs}. The model $\mathcal{G}$ has $N$ Gaussians, and each Gaussian is composed of the mean, $\mu$, rotation quaternion, $q$, scaling vector, $S$, and radar power return ratio, $\sigma$, which is the RCS value when a Gaussian represents a real object rather than noise: 
\begin{equation}
    \mathcal{G} = \{G_i:( \mu_i, q_i, S_i, \sigma_i)\ | i=1,...,N \}.
\end{equation}
Each radar frame $i$ has a pose $\mathbf{T}_i = [\mathbf{R}_i \,|\, \mathbf{t}_i] \in SE(3)$
that transforms world coordinates into the radar frame.

Given $\mathcal{G}$ and radar pose $\mathbf{T}_i$, the rendered radar range-azimuth image is obtained as:
\begin{equation}
\hat{\mathbf{I}}_{RA} = \Pi(\mathcal{G}, \mathbf{T}_i) \in \mathbb{R}^{N_r \times N_a},
\end{equation}
where $\Pi_{\text{radar}}$ is the differentiable radar renderer in \cite{radarsplat} based on radar equation (Eq.~\ref{eq:radar_equation}). Compared to the original GS~\cite{3dgs}, RadarSplat changes the sensor model to orthogonal weighted averaging in polar space along the elevation axis and renders based on the RCS $\sigma$ value.

\begin{figure}[t!]
    \centering
    \includegraphics[width=0.99\linewidth]{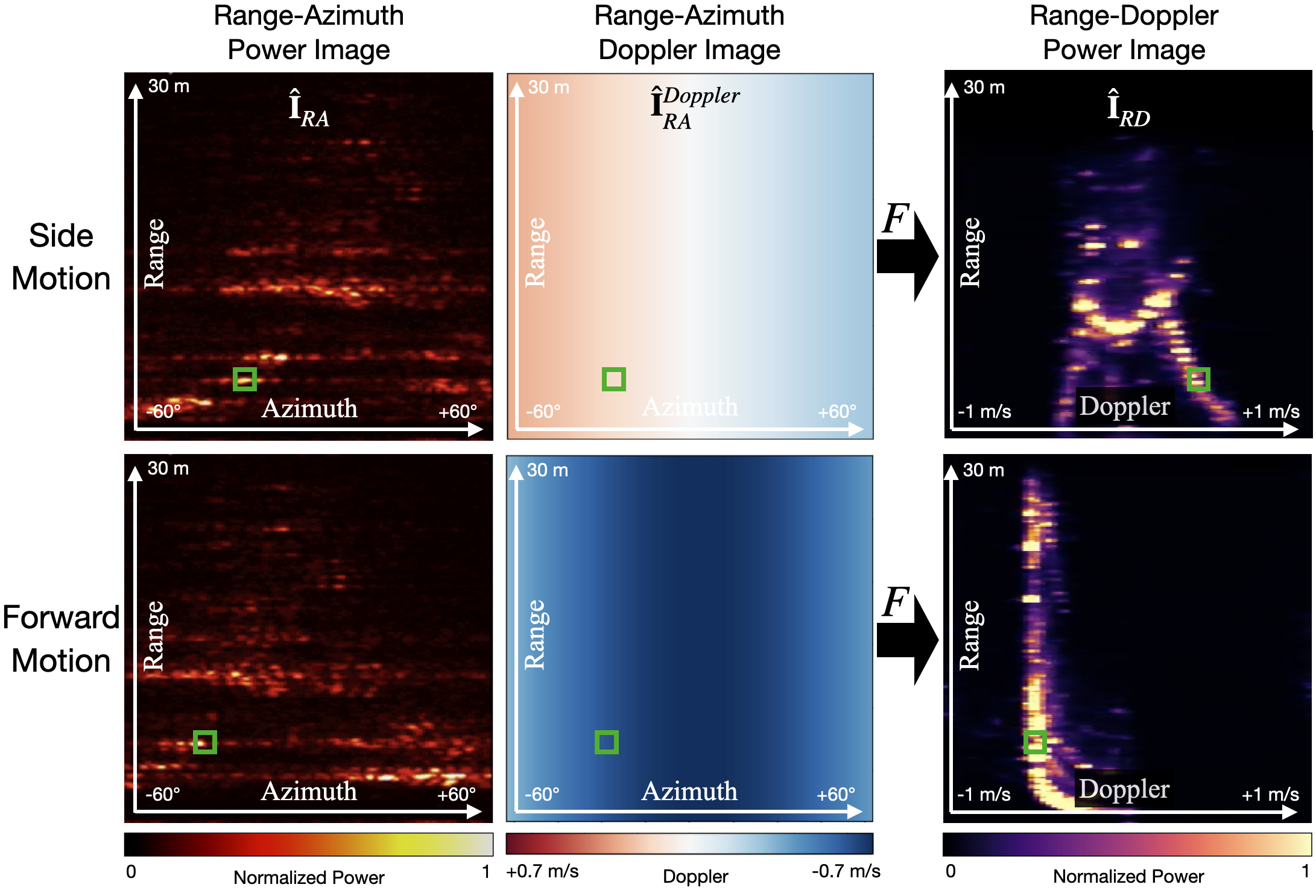}
    \vspace{-0.2in}
    \caption{Range–Doppler rendering examples. Range–Doppler images are rendered from the range–azimuth image and the Doppler map computed from ego-velocity. The first and second rows show side and forward motions, yielding distinct Doppler patterns. $F$ denotes the rendering operation. }
    \vspace{-0.1in}
    \label{fig:range_doppler_rendering}
\end{figure}

\begin{figure*}[]
    \centering
    \includegraphics[width=0.9\linewidth]{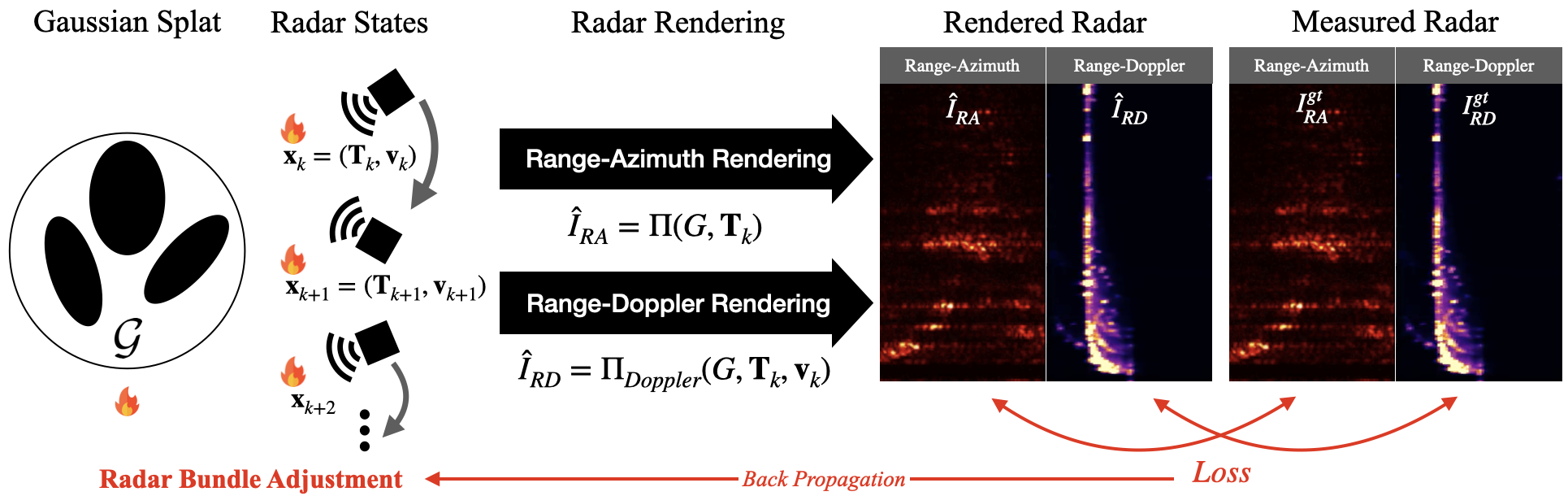}
    \vspace{-0.15in}
    \caption{Overview of RadarSplat++ rendering and training pipeline. Given radar poses and ego-velocity, RadarSplat renders both range–azimuth and range–Doppler images, which are compared with measured radar images to compute losses for scene reconstruction and pose optimization.}
    \label{fig:overview}
\end{figure*}

\subsubsection{Proposed Differentiable Range-Doppler Rendering}
MIMO Radar produces a data cube with spatial and velocity information. Unlike RadarSplat, which supports only RA rendering, we extend the framework to RD rendering, allowing Doppler velocity to contribute to both scene reconstruction and pose estimation, thereby fully exploiting the radar’s sensing capability.

To render radar range-Doppler image, we first compute the estimated Doppler velocity of each Gaussian based on ego-velocity $\mathbf{v}$ by assuming the measurement is static.
Given $\mathcal{G}$, radar pose $\mathbf{T}_i$, and radar velocity $\mathbf{v}$ obtained from $\mathbf{T}_k$ and $\mathbf{T}_{k-1}$ and time difference $\Delta k$, the Doppler velocity of a Gaussian is
\begin{equation}
v_{i}^{\text{Doppler}} = -\hat{\mathbf{r}}_i^\top \mathbf{v}
\end{equation}
, where
$\hat{\mathbf{r}}_i \;\triangleq\; 
\frac{\mu_i - \mathbf{t}}
{\lVert \mu_i - \mathbf{t} \rVert}$, and 
$\hat{\mathbf{r}}_i$ is the line-of-sight direction vector of each Gaussian.
The predicted radar range-Doppler image is obtained as:
\begin{equation}
\hat{\mathbf{I}}^{Doppler}_{RA}[r,a] = \Pi_{Doppler}(\mathbf{v}_i) = v_x \cos\theta_a + v_y \sin\theta_a 
\end{equation}
, where $\Pi_{Doppler}$ generate Doppler image based on ego-velocity $\mathbf{v}$ and $\theta_a$ is azimuth angle at bin $a$.
\begin{equation}
\hat{\mathbf{I}}_{RD} = F(\hat{\mathbf{I}}_{RA}, \hat{\mathbf{I}}_{RA}^{Doppler}, \Phi)  \in \mathbb{R}^{N_r \times N_d}
\end{equation}
, where $F$ converts the power and Doppler images in range-azimuth space, $\mathbf{I}_{RA}$ and $\mathbf{I}_{RA}^{\text{Doppler}}$, 
into a range-Doppler image weighted by the antenna profile $\Phi(\theta)$. The $F$ conversion is:

\begin{align}
\hat{\mathbf{I}}_{RD}[r,d]
&=\sum_{a}
\Big(\Phi(\theta_a)\,\hat{\mathbf{I}}_{RA}[r,a]\Big)\,
\nonumber\\
&\quad \cdot
\exp\!\left(
-\frac{\big(\hat{\mathbf{I}}^{\mathrm{Doppler}}_{RA}[r,a]-v_d\big)^2}{2\sigma^2}
\right)
\end{align}
, where $\Phi$ is radar antenna gain depend on $\theta_a$, $v_d$ is Doppler velocity at bin $d$, and $\sigma$ is the bandwidth of the soft binning Gaussian kernel. 
The design make sure $\hat{I}_{RD}$ is differentiable w.r.t. $\hat{I}_{RA}$ and $\mathbf{v}_i$.
In practice, we set $\sigma = 3\,\Delta v_d$, where $\Delta v_d$ denotes the Doppler-bin resolution, and use a small binning window $b=10$ to reduce memory and runtime overhead.
The rendering process is shown in Figure~\ref{fig:range_doppler_rendering}.

\subsection{Backend: RadarSplat++ Bundle Adjustment Optimization}
\label{sec:backend}
Unlike existing radar SLAM systems that lack joint multi-view and scene optimization, our backend performs radar BA through the RadarSplat++ (Sec~\ref{sec:radarsplat++}) scene representation inspired by NeRF/GS-based SLAM~\cite{gsslam, niceslam, loner}. The differentiable GS representation naturally supports BA joint optimization capability.
Following LONER~\cite{loner}, we deploy pose refinement and mapping modules before BA to obtain a better initial pose and map for the new keyframe.

\subsubsection{Pose Refinement} To refine a newly observed radar keyframe poses from frontend, we define the loss function to minimize rendering error from RA image $\mathbf{I}_{RA}$ and RD image $\mathbf{I}_{RD}$: 
\begin{multline}
\mathcal{L}_{pose} = \min_{\{\mathbf{T}_i \in W\}}
\sum_{i \in W}
\mathcal{L}
\left(
\hat{\mathbf{I}}_{RA}(\mathcal{G}, \mathbf{T}_i), \mathbf{I}_{RA}
\right)
+ \\
\mathcal{L}
\left(
\hat{\mathbf{I}}_{RD}(\mathcal{G}, \mathbf{T}_i, \mathbf{v}_i(\mathbf{T}_i)), \mathbf{I}_{RD}
\right)
,
\end{multline}
where the $\mathbf{v}_i$ is the ego-velocity estimation initialized by frontend RIO (Sec.~\ref{sec:rio}) and updated recurrently by estimate transformation $ \mathbf{T}_i$ throughout the optimization process. In practice, we only optimized the last keyframe, so the local window $W=\{KF_{k}\}$ when we have $k$ keyframes.

\subsubsection{Mapping} 
During mapping, only the geometry-matching loss from the RA image $\mathbf{I}_{RA}$ is used. An additional regularization term $\mathcal{L}_{\text{s}}(\boldsymbol{S})$ constrains the Gaussian size to prevent the formation of excessively large Gaussians.
\begin{multline}
\mathcal{L}_{map} = \min_{\mathcal{G}}
\sum_{i \in W}
\mathcal{L}
\left(
\hat{\mathbf{I}}_{RA}(\mathcal{G}, \mathbf{T}_i), \mathbf{I}_{RA}
\right)
+
\lambda_{\Sigma}\,\mathcal{L}_{\text{s}}(\boldsymbol{S})
,
\end{multline}
we use sliding window size with $N=10$ for the local mapping window $W=\{ KF_i \}_{i = k-N}^{k}$.

\subsubsection{Bundle Adjustment}
To jointly optimize radar poses and Gaussian parameters, we combine both pose and mapping optimizations. The BA loss term is defined as:
\begin{multline}
\mathcal{L}_{BA} = \min_{\mathcal{G},\, \{\mathbf{T}_i \in W\}}
\sum_{i \in W}
\mathcal{L}
\left(
\hat{\mathbf{I}}_{RA}(\mathcal{G}, \mathbf{T}_i), \mathbf{I}_{RA}
\right)
+ \\
\mathcal{L}
\left(
\hat{\mathbf{I}}_{RD}(\mathcal{G}, \mathbf{T}_i, \mathbf{v}_i(\mathbf{T}_i)), \mathbf{I}_{RD}
\right)
+
\lambda_{\Sigma}\,\mathcal{L}_{\text{s}}(\boldsymbol{S})
,
\end{multline}
where $\mathcal{L}$ is a combination of $L_1$ and SSIM-based reconstruction loss following~\cite{radarsplat}. The radar BA pipeline is shown in Figure~\ref{fig:overview}. We use radius window size with $r_{BA}=10~m $ for the BA window $W$.

\section{Experiments}

\begin{figure}[t!]
    \centering
    \includegraphics[width=0.9\linewidth]{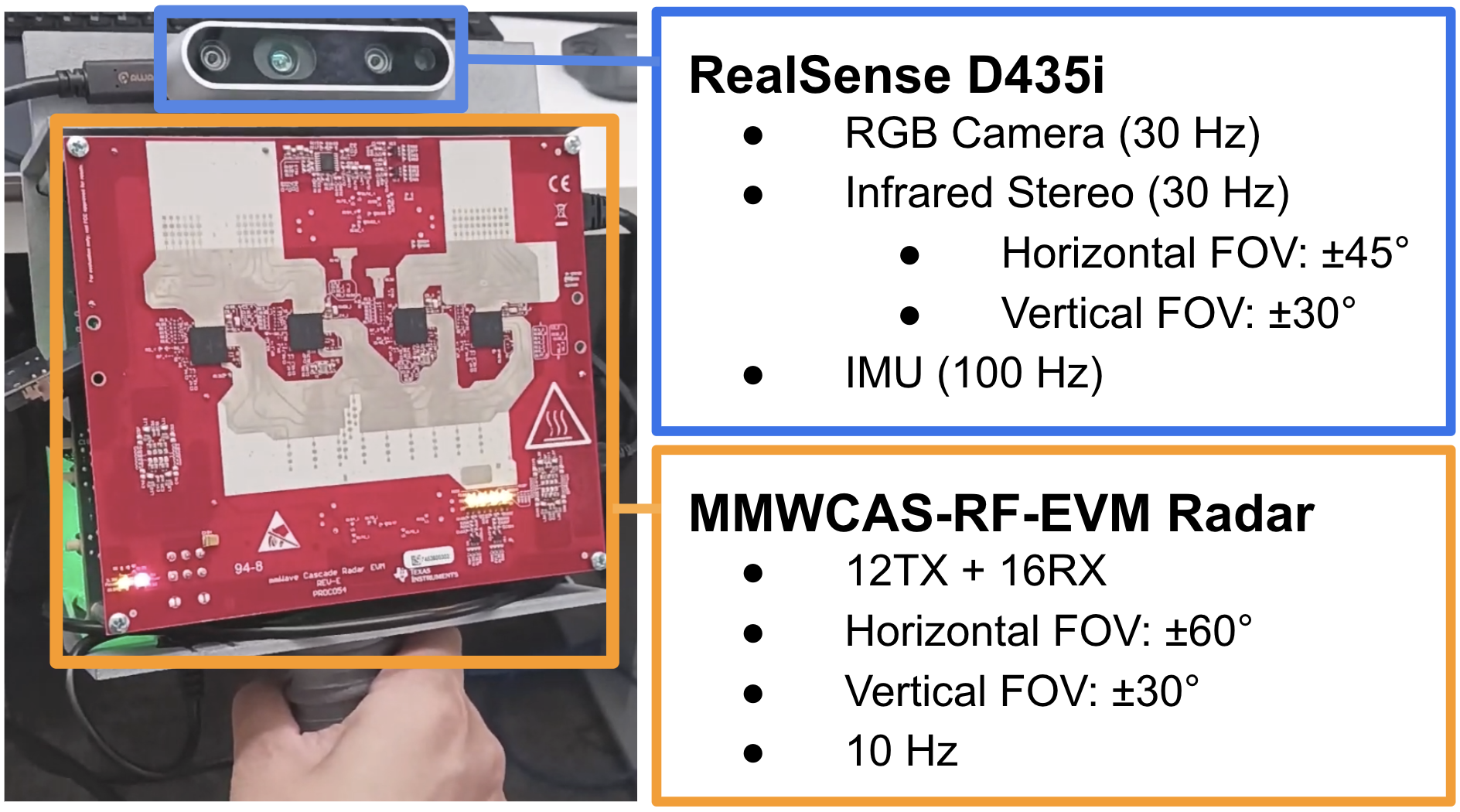}
        \vspace{-0.15in}
        \caption{Data collection sensor rig setup. }
        \label{fig:sensor_setup}
\end{figure}

\begin{figure}[t!]
    \centering
    \includegraphics[width=0.9\linewidth]{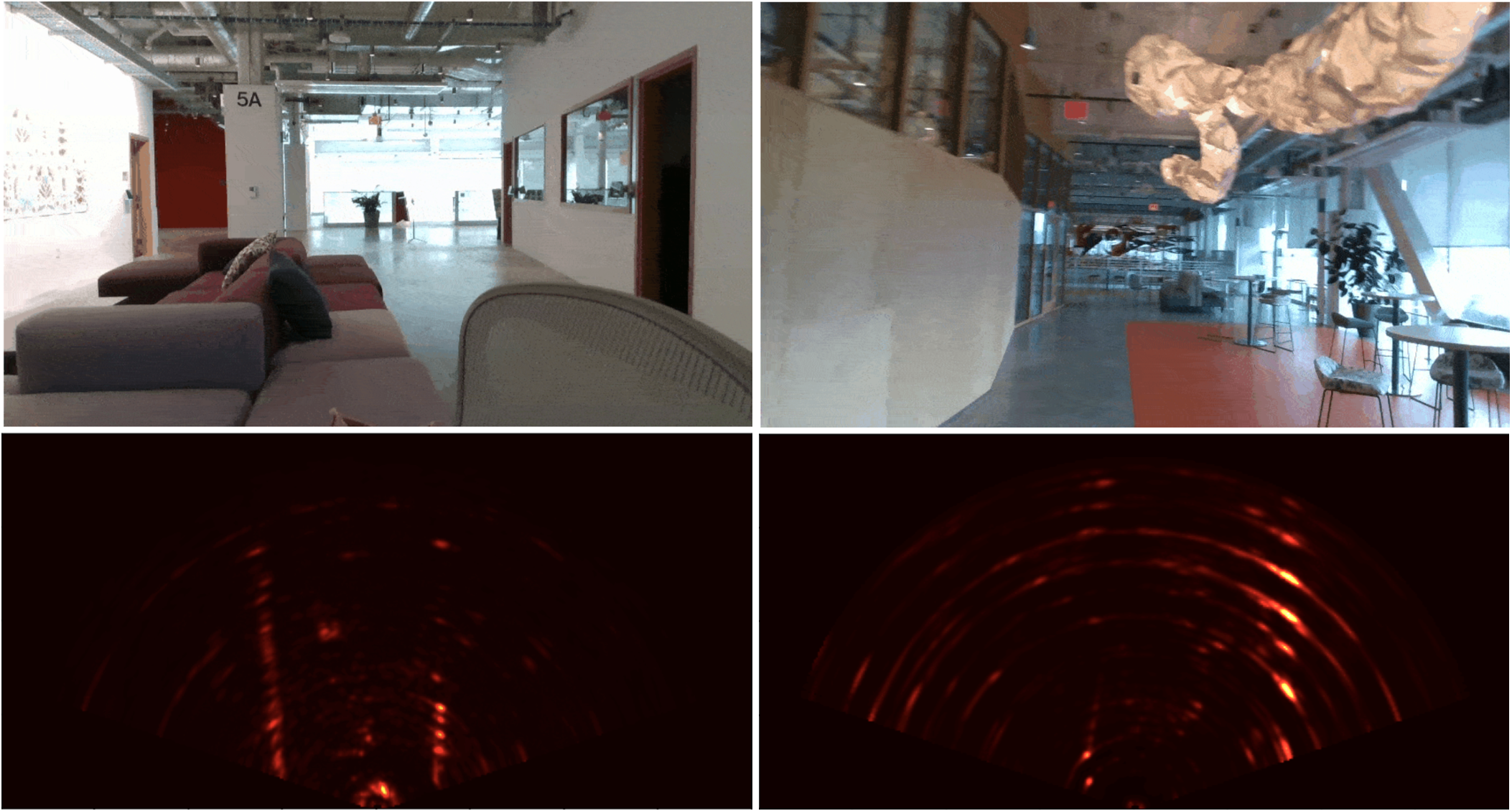}
        \vspace{-0.1in}
        \caption{Visualization of camera and radar data from our indoor dataset. The radar range–azimuth images in Cartesian space are shown to provide a clearer sense of the radar observations.}
        \label{fig:collected_data}
\end{figure}

\begin{table}[!t]
\centering
\renewcommand{\arraystretch}{1.2}
\setlength{\tabcolsep}{6pt}
\resizebox{\linewidth}{!}{%
\begin{tabular}{llcc}
\hline
Name       & Description      & Travel Distance (m) & Length (s) \\ \hline
Sequence 1 & Meeting Room     & 42.46               & 141.07    \\
Sequence 2 & Small Loop       & 91.89               & 157.98    \\
Sequence 3 & Large Loop       & 144.59              & 232.74    \\
Sequence 4 & Small + Large Loop & 206.12              & 263.19    \\
Sequence 5 & Medium + Large Loop   & 233.54              & 320.06    \\ \hline
\end{tabular}
}
\vspace{-0.05in}
\caption{Experiment sequences details.
}
\vspace{-0.25in}
\label{tab:sequences}
\end{table}

\begin{table*}[!h]
\centering
\renewcommand{\arraystretch}{1.3}
\setlength{\tabcolsep}{9pt}
\begin{tabular}{lcccccccccc}
\hline
\multicolumn{1}{c}{\multirow{2}{*}{\textbf{Scenes}}} & \multicolumn{2}{c}{\textbf{Sequence 1}} & \multicolumn{2}{c}{\textbf{Sequence 2}} & \multicolumn{2}{c}{\textbf{Sequence 3}} & \multicolumn{2}{c}{\textbf{Sequence 4}} & \multicolumn{2}{c}{\textbf{Sequence 5}} \\ \cline{2-11} 
\multicolumn{1}{c}{}                                 & \textit{MRIO}       & Ours               & \textit{MRIO}       & Ours               & \textit{MRIO}       & Ours               & \textit{MRIO}       & Ours               & \textit{MRIO}       & Ours               \\ \hline
\textbf{Trans. (m)}                                  & 0.94               & \textbf{0.19}      & 7.57               & \textbf{0.42}      & 22.48              & \textbf{2.04}      & 11.04              & \textbf{1.05}      & 12.9               & \textbf{0.97}      \\
\textbf{Rot. ($^\circ$)}                                  & 5.03               & \textbf{1.91}      & 15.51              & \textbf{2.06}      & 32.38              & \textbf{5.25}      & 17.62              & \textbf{2.66}      & 21.55              & \textbf{3.28}      \\ \hline
\end{tabular}
\caption{Absolute Pose Error comparison between MRIO~\cite{mrio} and our method on various scenes.}
\vspace{-0.05in}
\label{tab:ape_comparison}
\end{table*}

\begin{figure*}[t!]
    \centering
    \includegraphics[width=0.99\linewidth]{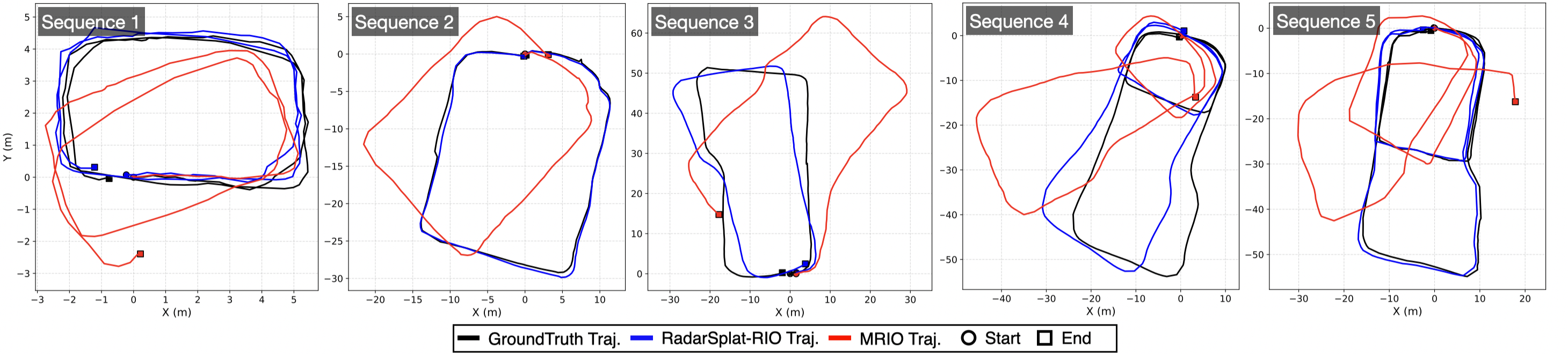}
        \vspace{-0.1in}
        \caption{Comparison of estimated trajectory between MRIO~\cite{mrio} and proposed method.}
        \vspace{-0.05in}
        \label{fig:traj}
\end{figure*}
\begin{figure}[t!]
    \centering
    \includegraphics[width=0.9\linewidth]{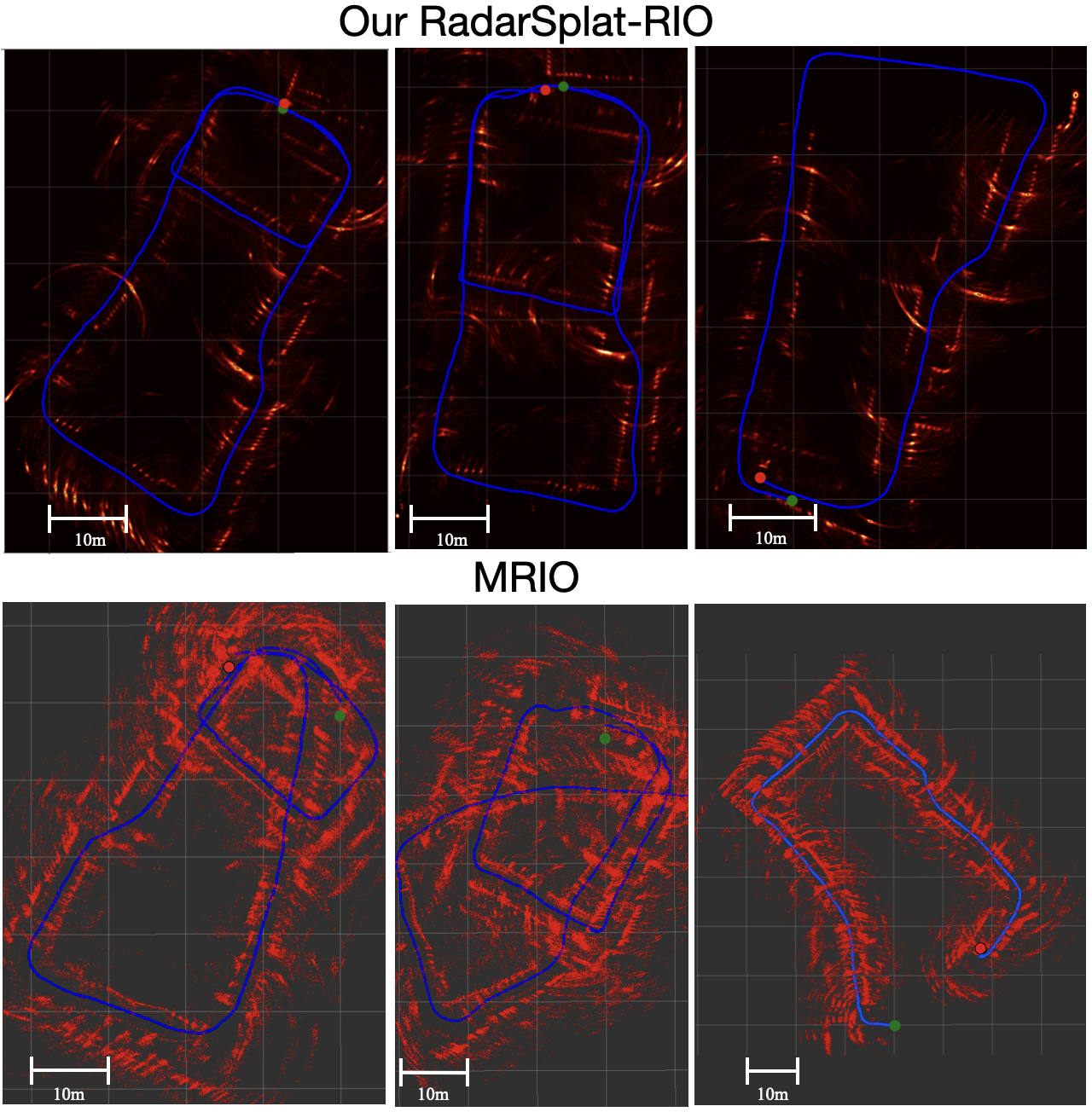}
        \caption{Scene reconstruction comparison between MRIO and the proposed RadarSplat-RIO. The proposed method not only greatly reduces odometry drift but also produces a dense and consistent scene reconstruction.}
        \label{fig:recon}
\end{figure}

\subsection{Dataset}
Existing radar SLAM datasets typically provide only partial radar representations, such as RA images~\cite{oxford}, RD images~\cite{dart}, or radar point clouds~\cite{nuscenes,vod,xrio}. Among publicly available datasets, ColoRadar~\cite{coloradar} is the only one that provides raw RAD data. However, we found that the released Doppler data are not compatible with the existing RIO method~\cite{mrio} designed for multi-chip radar.
Therefore, we evaluate the proposed method on a custom dataset collected using a TI MMWCAS-RF-EVM millimeter-wave radar and an Intel RealSense D435i RGB-D camera (Figure~\ref{fig:sensor_setup}) following data collection pipeline in \cite{mrio}. Following MRIO~\cite{mrio} and X-RIO~\cite{xrio}, we adopt stereo VIO as pseudo-ground truth, as LiDAR SLAM may fail in indoor monotone corridors, as discussed in~\cite{mrio}.
The RealSense D435i provides synchronized IMU measurements and supports off-the-shelf visual-inertial SLAM via RTAB-Map~\cite{rtabmap}. We use RTAB-Map to obtain reference trajectories, which serve as pseudo ground truth for evaluation.
The dataset consists of five indoor sequences captured under varying motion patterns. Figure~\ref{fig:collected_data} shows camera-radar pairs of the collected data.
Each sequence contains at least one, and some contain two, loop closures of different sizes, allowing clear observation of trajectory drift and its correction through the proposed BA. The sequences are summarized in Table~\ref{tab:sequences}. The dataset will be publicly released upon paper acceptance.

\subsection{Baseline}
While numerous radar SLAM methods exist, they are often tailored to specific radar types and require significant tuning to generalize. To the best of our knowledge, X-RIO~\cite{xrio} and Radarize~\cite{radarize} are the only open-source methods using a similar TI radar setup. However, MRIO~\cite{mrio} reports that X-RIO is designed for single-chip radar and does not transfer to multi-chip radar, which we also observe. Radarize shows similarly unstable performance in our experiments.
Therefore, we compare our method with MRIO~\cite{mrio}, designed for the same multi-chip TI radar used in this study. We also adopt MRIO as the frontend to isolate the contribution of the proposed RadarSplat++ backend BA.


\subsection{Odometry Evaluation}
In Table~\ref{tab:ape_comparison}, we demonstrate that the proposed RadarSplat-RIO with BA backend significantly reduces the drifting issue. Figure~\ref{fig:traj} presents the qualitative comparison between MRIO and RadarSplat-RIO. 
The proposed method significantly improves the robustness of pose estimation and effectively mitigates long-term drift. The average absolute translational and rotational errors are reduced by more than 90\% and 80\%, respectively.
Figure~\ref{fig:recon} further shows that our method achieves superior scene reconstruction, enabled by more accurate radar pose estimation and dense scene mapping through the Gaussian scene representation. We note that the MRIO accuracy may be affected by imperfect time synchronization between the radar and the IMU embedded in the RealSense camera in our dataset due to hardware restrictions. However, this effect further highlights the robustness of the proposed backend optimization.

\begin{table}[!b]
\centering
\renewcommand{\arraystretch}{1.3}
\setlength{\tabcolsep}{6pt}
\resizebox{\linewidth}{!}{%
\begin{tabular}{clclclcc}
\hline
\multicolumn{8}{c}{Method Components Ablations}                                                                              \\ \hline
Frontend                  &  & \multicolumn{3}{c}{Backend}                              &  & \multicolumn{2}{c}{ATE}         \\ \cline{1-1} \cline{3-5}
RIO                       &  & Loc. \& Map.              &  & BA                        &  & Trans. (m)     & Rot ($^\circ$) \\ \cline{1-5} \cline{7-8} 
\checkmark &  & \checkmark &  & \checkmark &  & \textbf{0.934} & \textbf{3.03}  \\
\checkmark &  &                           &  &                           &  & 10.99          & 18.42          \\
\checkmark &  & \checkmark &  &                           &  & 2.83           & 8.23           \\
                          &  & \checkmark &  & \checkmark &  & 18.27          & 25.91          \\ \hline
\end{tabular}
}
\vspace{-0.1in}
\caption{
Ablation study comparing the proposed full method, 
without the proposed backend, 
without the proposed BA, 
and without RIO frontend across all sequences. 
Metrics are Absolute Pose Error (APE) in translation (m) and rotation (deg). Lower is better.
}
\vspace{-0.25in}
\label{tab:ablation_method}
\end{table}

\begin{table*}[!h]
\centering
\renewcommand{\arraystretch}{1.2}
\setlength{\tabcolsep}{9pt}
\begin{tabular}{ccccccccccccc}
\hline
\multirow{3}{*}{Ablations} &  & \multicolumn{8}{c}{BA Frames Selection Strategy}                                           &  & \multicolumn{2}{c}{RadarSplat++}  \\ \cline{3-10}
                           &  & \multicolumn{3}{c}{Radius Spatial Window} &  & \multicolumn{4}{c}{Sliding Window} &  & \multicolumn{2}{c}{RD-image Loss} \\ \cline{3-5} \cline{7-10} \cline{12-13} 
                           &  & $r_{BA}=1$ & $r_{BA}=5$ & $r_{BA}=10~(\text{Selected})$  &  & N=2   & N=5   & N=10  & N=$\infty$ &  & w/                   & w/o        \\ \hline
\textbf{Trans. (m)}        &  & 3.12        & 1.07        & \textbf{0.93} &  & 2.83  & 1.42  & 1.17  & 1.36       &  & \textbf{0.93}        & 3.41       \\
\textbf{Rot. ($^\circ$)}   &  & 7.98        & 3.24        & \textbf{3.03} &  & 8.23  & 3.95  & 3.61  & 4.03       &  & \textbf{3.03}        & 3.57       \\ \hline
\end{tabular}
\caption{
Ablation study comparing different BA window selection strategies and the RadarSplat++ Range-Doppler image loss. 
Metrics are Absolute Pose Error in translation (m) and rotation (deg). Lower is better.
\vspace{-0.2in}
}
\label{tab:ablation_window}
\end{table*}

\subsection{Ablation Studies}
\subsubsection{Method Components}
In Table~\ref{tab:ablation_method}, we evaluate the impact of the main components of our pipeline by disabling each one individually. Removing the entire RadarSplat-RIO backend causes the trajectory to drift rapidly, as the system relies solely on RIO and therefore accumulates substantial drift. Disabling the BA module in the backend also degrades accuracy, consistent with prior observations~\cite{balm, loner} that BA is essential for precise pose estimation. Finally, omitting RIO as the initial pose estimator results in the poorest performance due to the lack of a reliable pose initialization for the optimizer.

\subsubsection{BA Keyframe Selection Strategy}
In Table~\ref{tab:ablation_window}, we show that the selected radius-based optimization window with $r_{BA}=10,\mathrm{m}$ outperforms configurations using different values of $r_{BA}$ as well as temporal sliding-window bundle adjustment (BA). The radius-based strategy selects spatially relevant keyframes when the trajectory revisits previously mapped regions, improving geometric consistency without requiring explicit loop-closure detection.

\subsubsection{RadarSplat++ Range-Doppler Rendering}
In Table~\ref{tab:ablation_window}, we further compare performance with and without the RadarSplat++ RD rendering loss. The RD image loss primarily improves translation accuracy, as Doppler measurements directly constrain linear motion. We observe that the benefit is most pronounced in featureless corridors, where RA-only image alignment becomes ill-conditioned.

\section{Conclusion}
We have introduced RadarSplat-RIO, the first radar odometry framework that incorporates bundle adjustment using a GS scene representation, enabling accurate and drift-resilient odometry. Unlike prior works, which only use RA~\cite{radarsplat} or RD~\cite{dart} images for radar NeRF/GS reconstruction. The proposed RadarSplat++ extends the RadarSplat~\cite{radarsplat} formulation to RD image rendering, which enables the full power of radar RAD data for pose estimation and scene reconstruction. Our method improves upon prior RIO method by mitigating drift without relying on explicit loop closure. 

The experiments demonstrate that our approach substantially improves trajectory accuracy and robustness, reducing average absolute translational and rotational errors by more than 90\% and 80\%, respectively. This highlights the effectiveness of our radar BA. Although we evaluate the system using a single RIO frontend, the proposed RadarSplat-RIO backend can be integrated with various radar or RIO frontends. We believe this work lays the foundation for accurate and reliable radar SLAM, paving the way for its deployment in robotics and autonomous navigation under challenging conditions. The current method is limited to single-radar 3-DoF pose estimation as a proof of concept. In future work, we aim to generalize the framework to multi-radar 6-DoF pose estimation following~\cite{xrio,mrio}.

\fontsize{8.5pt}{10pt}\selectfont
\bibliographystyle{IEEEtran}
\bibliography{main}

\end{document}